\documentclass{article}
\usepackage{amsmath,graphicx,mlspconf}
\usepackage[colorlinks=true, linkcolor=blue, citecolor=blue, urlcolor=blue]{hyperref}
\usepackage{xcolor}
\usepackage{booktabs}
\usepackage{markdown}
\usepackage{graphicx}
\usepackage[style=ieee, maxnames=2, minnames=1]{biblatex}
\copyrightnotice{979-8-3503-2411-2/25/\$31.00 {\copyright}2025 IEEE}

\toappear{2025 IEEE International Workshop on Machine Learning for Signal Processing, Aug.\ 31-- Sep.\ 3, 2025, Istanbul, Turkey}

\title{Trust the Model: Compact VLMs as In-Context Judges for Image-Text Data Quality}
\name{%
\begin{tabular}{c}
   Daulet Toibazar \thanks{Corresponding author: dtoibazar@humain.ai} \qquad
   Kesen Wang \qquad
   Sherif Mohamed \qquad \\
   Abdulaziz Al-Badawi \qquad
   Abdulrahman Alfulayt \qquad
   Pedro J. Moreno%
\end{tabular}
}

\address{
Humain\\ Riyadh, KSA \\
}

\begin{document}
%\ninept

\maketitle

\begin{abstract}
Vision-language models (VLMs) extend the conventional large language models by integrating visual data, enabling richer multimodal reasoning and significantly broadens the practical applications of AI. However, including visual inputs also brings new challenges in maintaining data quality. Empirical evidence consistently shows that carefully curated and representative training examples often yield superior results compared to simply increasing the quantity of data. Inspired by this observation, we introduce a streamlined data filtration framework that employs a compact VLM, fine-tuned on a high-quality image-caption annotated dataset. This model effectively evaluates and filters potential training samples based on caption and image quality and alignment. Unlike previous approaches, which typically add auxiliary filtration modules on top of existing full-scale VLMs, our method exclusively utilizes the inherent evaluative capability of a purpose-built small VLM. This strategy eliminates the need for extra modules and reduces training overhead. Our lightweight model efficiently filters out inaccurate, noisy web data, improving image-text alignment and caption linguistic fluency. Experimental results show that datasets underwent high-precision filtration using our compact VLM perform on par with, or even surpass, larger and noisier datasets gathered through high-volume web crawling. Thus, our method provides a lightweight yet robust solution for building high-quality vision-language training corpora. \\ \textbf{Availability and implementation:} Our compact VLM filtration model, training data, utility scripts, and Supplementary data (Appendices) are freely available at \url{https://github.com/daulettoibazar/Compact_VLM_Filter}.

\end{abstract}

\newcommand{\cem}[1]{\textcolor{blue}{cem: #1}}
\section{Introduction}
\label{sec:intro}

Vision-Language Models (VLMs) \cite{gemini} \cite{qwen2vl} \cite{gpt4} \cite{Qwen25_vl} \cite{llava_one_vision} \cite{deepseekvl} have significantly advanced the field of AI by incorporating visual signals into large language model (LLM) reasoning, enabling multimodal understanding and improving performance in tasks such as image captioning, visual question answering, and document comprehension. These advancements have been largely driven by recent innovations in model architectures \cite{qwen2vl} \cite{alignvlm} \cite{global_semantic_feature_allocation} \cite{mplug_owl}, alongside the availability of large-scale and diverse training datasets \cite{cc12m} \cite{recap_datacomp_1B} \cite{obelics} \cite{lgs}. To train VLMs to effectively follow user instructions and exhibit strong visual reasoning capabilities, a combination of techniques remains essential—including visual alignment during pre-training, supervised fine-tuning (SFT), and reinforcement learning with human feedback (RLHF) \cite{alignvlm} \cite{qformer} \cite{rlhf} \cite{llava1.5}. However, the quality of training data has a direct impact on how well these models generalize to real-world tasks. Recent research has demonstrated that smaller, meticulously selected datasets frequently perform better than much larger but noisier corpora, underscoring the crucial role that data refinement plays in VLM training \cite{lima} \cite{influencenet} \cite{selffilter}.

Publicly available multimodal datasets, typically compiled through extensive web scraping, contain various forms of noise, including irrelevant or incorrect captions (see Appendix A) and inherited biases. Training VLMs on such noisy data risks substantial degradation in model performance, e.g.,  manifested in hallucinations \cite{survey_hallucination} \cite{origin_of_hallucination} and limits their performance in visual tasks. In response, numerous dataset curation strategies have been proposed \cite{selffilter} \cite{instruction_gpt4}, with recent methods increasingly utilizing VLMs for annotation and quality evaluation, a practice known as model-based filtration. Leveraging the sophisticated interpretive capabilities of these models, our work introduces a simple yet effective approach for data quality assessment: fine-tuning a compact VLM on the carefully annotated, high-quality dataset, enabling it to act as an efficient and interpretable scoring function. 

\begin{figure}[htbp]
\centering
\includegraphics[width=0.48\textwidth]{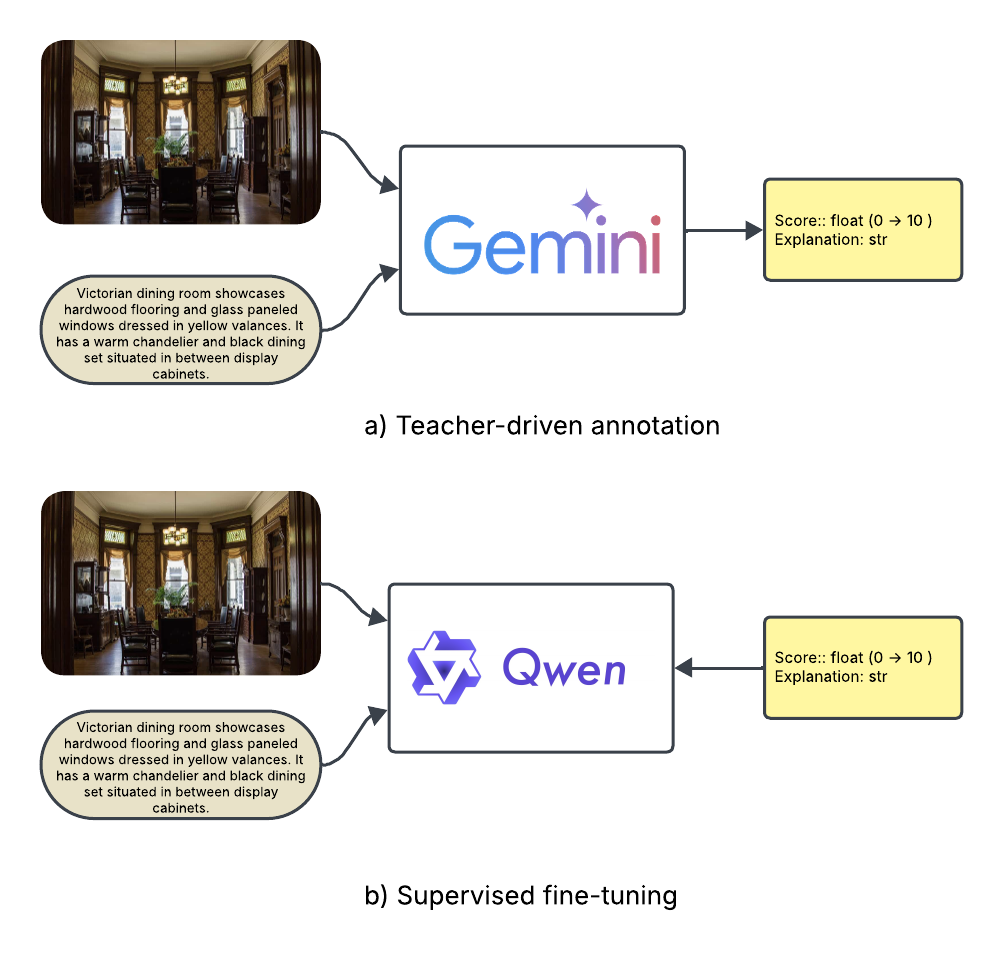}
\caption{Two‐stage framework for automated visual–textual quality assessment. a) The teacher-driven annotation pipeline employs Gemini as a powerful teacher model to assign quality scores (1–10) and rationales to image–text pairs sourced from public caption datasets. b) Supervised fine‐tuning the compact Qwen2VL model, resulting in a lightweight, efficient scoring function.\label{fig1}}

\end{figure}

Our approach is cost- and resource-effective than previous approaches like \cite{alpagasus}, which rely on external API calls. The compact size of our VLM filtration model allows on-prem deployment and filtration of huge image-caption data. We also leverage the inherent multimodal understanding capabilities of VLMs, rather than relying on CLIP- or FFN-based scoring functions, which are frequently limited in interpretability and semantic richness. Our proposed model uses specific metrics described in Appendix B to evaluate the image-caption quality. An overview of our suggested approach is shown in Figure~\ref{fig1}.

Our contributions are summarized as follows: 
\begin{itemize} 
    \item We curate and release the dataset used to train a filtration model for assessing the semantic alignment of image-text pairs.
    \item Through empirical research, we confirm that it is feasible to fine-tune a compact VLM to effectively filter multimodal data. 
    \item Our lightweight scoring VLM operates independently of external tools, making it accessible and free for community use. 
\end{itemize}

\section{Related works}
\label{sec:litreview}

Recent studies have highlighted the critical importance of selective data curation in training VLMs. For instance, LIMA \cite{lima} showed that a smaller, highly curated instruction dataset outperforms larger and noisier datasets. However, the manual prompt selection approach used there limits broader applicability across diverse multimodal domains. InstructionGPT-4 \cite{instruction_gpt4} further improves by fine-tuning MiniGPT-4 \cite{minigpt4} with a learnable data selector, achieving strong results with only 6\% of the original dataset. However, its reliance on multiple scoring signals introduces interpretation complexity. Similarly, Self-Filter \cite{selffilter} utilized a co-trained VLM-based scoring network to efficiently identify challenging multimodal training samples, achieving results comparable to full-dataset training with just 15\% of the data. 

Building on these insights, we introduce a lightweight and interpretable filtration model designed to evaluate and filter image–caption data in terms of their alignment and quality (see Appendix B). Our model was fine-tuned on supervision signals derived from a substantially larger and more capable teacher model, which provides both a scalar filtration score and an explanatory rationale behind the score. Our approach focuses on the pre-training stage of VLMs, where image-text pair quality is crucial but frequently neglected. We aim to enhance the integrity of training data by implementing selective filtering at this fundamental level.

\section{Methodology}
\label{sec:methods}
We ensured high-quality supervision by using the state-of-the-art multimodal model Gemini 2.0-Flash \cite{gemini} to annotate our training data. Each data point received a numerical quality score (1–10) and a detailed textual explanation, allowing the downstream filtration model to learn both how to score and explain image-caption quality. To address potential errors from automated annotations, we manually reviewed annotated data to ensure accuracy and reliability.

We then applied supervised fine-tuning (SFT) to a compact vision–language model, Qwen2-VL-2B \cite{qwen2vl}, training it on the Gemini-generated annotations. This allowed the smaller model to approximate the teacher’s scoring behavior and explanations. The fine-tuned model offers a cost- and compute-efficient solution for filtering large-scale multimodal datasets, without depending on expensive APIs or large models.

\subsection{Data annotation}
\label{sec:data_annotation}

\begin{table*}[htbp]
\caption{Overview of datasets used to train our compact VLM filtration model \label{table1}}
\centering
\begin{tabular}{lllll}
\toprule
\textbf{Dataset} & \textbf{Original size} & \textbf{Selected samples} & \textbf{Source} & \textbf{Avg. caption length} \\
\midrule
CC12M        & 12M   & 2.5K  & Web-scale image-text pairs & 18 words \\
Recap-COCO   & 30K   & 2.5K  & DataComp-1B                & 55 words \\
\bottomrule
\end{tabular}
\end{table*}

We used two commonly adopted datasets for pre-training vision–language models: Recap-COCO \cite{recap_datacomp_1B}, known for its high-quality re-captioned data, and CC12M \cite{cc12m}, a large-scale collection of web-scraped image-caption pairs. From these, we selected a balanced subset of 5,000 representative samples (see Table~\ref{table1}). By combining high-quality and noisy captions, we expose the filtration model to a broad range of data quality, enabling it to learn fine-grained distinctions in image-text alignment and improve generalization.

We prompted the teacher model to assign each pair a continuous quality score from 1 (lowest) to 10 (highest), along with a textual explanation justifying the score. This provides a richer training signal than simple binary labels. To ensure the accuracy of the annotations, all samples were manually reviewed after the initial automated labeling.

\subsection{Training filtration model}
\label{sec:train_filtration}
We used the annotated dataset to fine-tune our compact VLM (Qwen2VL-2B). By training this lightweight model with standard supervised fine-tuning, we enabled it to evaluate semantic alignment and safety considerations in image-caption pairs. Training was conducted on a GPU cluster with eight NVIDIA H100 GPUs, with a learning rate (lr) of $2 \times 10^{-6}$ and a cosine learning rate scheduler. We used a batch size of 128 and ran the training for a single epoch.

\subsection{Analyzing the impact of data filtration}
\label{sec:downstream_exp}

We evaluated the efficiency of our approach by filtering over 20K image-caption pairs: 10K from CC3M \cite{cc3m} and 10K from CC12M \cite{cc12m}, ensuring no overlap with the training data described in Section~\ref{sec:data_annotation}.

We retained only high-quality pairs—those receiving a score of 9 or higher from our compact VLM model—resulting in a filtered set of 3.5K pairs (\texttt{filtered}) from the original 20K (\texttt{full}). To control for dataset size effects, we also created a baseline \texttt{random} set by sampling 3.5K pairs randomly from the full dataset (matching the size of \texttt{filtered}, ~18\%).

We evaluated the quality of each set using the following metrics:

\begin{itemize}
\item Semantic alignment: We computed the mean cosine similarity (see Appendix C) between image and caption embeddings using the CLIP encoder \cite{clip}, providing a quantitative measure of semantic consistency.

\item Linguistic fluency and complexity: We calculated language model perplexity scores on each caption set (see Appendix D). Lower perplexity indicates more fluent and coherent captions, allowing comparison across filtered and unfiltered data.
\end{itemize}

\subsection{Downstream captioning performance}

To further validate our approach, we fine-tuned a lightweight image captioning model that connects a ViT-based vision encoder \cite{vit} to a GPT-2 decoder \cite{gpt2}, following prior work \cite{vit_gpt_1, vit_gpt_2}. We trained two versions of the model: one using the \texttt{filtered} dataset and another using the \texttt{full} dataset. Evaluation was performed on a consistent test set of 500 high-quality, previously unseen image-caption pairs.

To assess caption quality, we used an "LLM-as-a-judge" strategy, employing Gemini 2.0 Flash \cite{gemini} as the evaluator. Gemini was prompted to compare two captions—one from each model—against the ground-truth caption and identify which better matched it. The final score was computed as the percentage of cases in which captions from the \texttt{filtered} model were preferred over those from the \texttt{full} model.

\section{Analysis}
\label{sec:analysis}

\subsection{CLIP-based image-caption alignment}

Cosine similarity analysis on CLIP embeddings reveals that the \texttt{filtered} dataset exhibits stronger text-to-image alignment compared to both \texttt{random} and \texttt{full} datasets. This suggests that our filtration model effectively retains samples where the visual and textual modalities are more semantically aligned, i.e., the caption closely reflects the image content. This statement is reflected in the mean cosine similarity scores observed across the splits:
\[
\bar{S}_{\mathrm{full}} = 0.298,\quad
\bar{S}_{\mathrm{random}} = 0.297,\quad
\bar{S}_{\mathrm{filtered}} = 0.313.
\]

To quantitatively assess whether the observed improvement in cosine similarity is statistically meaningful, we conducted an independent two-sample Student’s \(t\)\nobreakdash-test using \texttt{SciPy}~\cite{scipy} library. Specifically, we tested whether the \texttt{filtered} set has a significantly higher mean cosine similarity than the \texttt{random} set:
\[
t = 15.87,\quad p = 4.27\times10^{-56}.
\]

This extremely low \(p\)\nobreakdash-value strongly rejects the null hypothesis of equal means in favor of the alternative, indicating that the improvement in cosine similarity for the \texttt{filtered} set is statistically significant.

\subsection{Caption quality boosts predictive performance}

The observed perplexity values across three splits are as follows:
\[
\mathrm{PPL}_{\mathrm{full}} = 170.2\quad
\mathrm{PPL}_{\mathrm{random}} = 168.6\quad
\mathrm{PPL}_{\mathrm{filtered}} = 137.2
\]
The \texttt{filtered} set exhibits a perplexity of 137.2, which is substantially lower than both \texttt{full} (170.2) and the \texttt{random} (168.6) sets.

This experiment yields 2 main findings: 

\begin{itemize}
    \item The lower perplexity of the \texttt{filtered} set demonstrates that our filtering procedure effectively removes noisy or misaligned captions and leaves behind cleaner, better-quality data. The fact that the \texttt{random} set of equal size is worse in perplexity confirms that the benefit is due to filtering quality and not dataset size reduction.

    \item Low perplexity means that models trained on the \texttt{filtered} captions benefit in generalizing better and avoiding overfitting or memorization of wrong patterns. This means our filtered dataset is not just better aligned with images but also benefits downstream training.
\end{itemize}

\subsection{Captioning performance}
To quantify the effect of our filtration strategy on caption quality, we fine-tuned identical ViT–GPT2 captioning models on both full and filtered datasets and generated captions for a set of 500 previously unseen images. Using Gemini 2.0 Flash as an automated judge, each ground-truth caption was paired with two model outputs; the LLM then selected which generated caption more closely matched the reference. Strikingly, in 59.4\% of cases (297/500), the judge favored the filtered model’s caption over that of the full model.  We interpret the 59.4\% preference rate as evidence that filtering alters the training signal in a measurable way. Notably, the filtered dataset constitutes only ~18\% of the original data, which inherently reduces the variance in the training data. Focused curation appears to shift model attention toward features and patterns that more closely align with caption quality, providing more accurate training signals. 

At the same time, the fact that the full-dataset model remains preferred in over 40\% of cases highlights that aggressive filtering may also discard examples that contribute valuable diversity. Together, these results suggest a nuanced trade-off in data curation: selective exclusion can improve alignment with reference captions under certain conditions but may also eliminate informative variability, pointing toward avenues for more refined or adaptive filtering strategies in future work.

\section{Ablation study}
\label{sec:ablation}

\begin{table}[htbp]
\caption{We sampled 4,994 image-text pairs from the CC12M dataset and assigned filtration scores using our compact filtration-oriented VLM. The table reports the average image-text alignment score (↑) across different score buckets, showing that higher filtration scores correspond to stronger semantic alignment. \label{table2}}
\resizebox{\linewidth}{!}{%
\begin{tabular}{lllll}
\toprule
\textbf{Score Range} & \textbf{1–3} & \textbf{4–6} & \textbf{7–8} & \textbf{9–10} \\
\midrule
Number of Samples & 577 & 246 & 3,238 & 933 \\
Alignment Score (↑) & 0.27 & 0.29 & 0.30 & 0.32 \\
\bottomrule
\end{tabular}%
}
\vspace{5 mm}
\footnotesize{\textit{Note:} ↑ indicates higher is better.}
\end{table}

We carried out an additional ablation study using 4,994 samples from the CC12M dataset to validate the effectiveness of our compact filtration model in assigning meaningful scores. Each sample was processed through the filtration model, where we kept all samples and their corresponding scores, ranging from 1 to 10 based on perceived alignment between the image and its accompanying text.

To analyze whether these scores truly reflect alignment quality, we grouped the samples into four score-based buckets: 1–3, 4–6, 7–8, and 9–10. For each bucket, we performed an image-text alignment evaluation using CLIP embeddings by computing the cosine similarity between image and text features. The results are summarized in Table \ref{table2}. Higher filtration scores correspond to higher similarity scores between image and text, confirming that our filtration model effectively captures their semantic alignment. The steady increase in alignment scores across buckets demonstrates that the model's scoring mechanism can serve as a reliable proxy for selecting good-quality image-text pairs, an essential step for improving downstream training quality.

\section{Conclusion}
\label{sec:conclusion}
Our research shows that data curation on image-caption datasets yields noticeable gains in a number of evaluation metrics. We created a \texttt{filtered} dataset that improves caption coherence in addition to downstream captioning performance. Language model perplexity significantly decreased in the filtered data, suggesting that improved predictive modeling is made possible by more precise and contextually relevant captions. Our data curation approach leads to stronger image-text alignment, quantified via higher CLIP-based cosine similarity. The statistical significance of the similarity gains, supported by an extremely low \(p\)\nobreakdash-value, confirms that semantic alignment is not incidental. 

Complementing the quantitative results, the Gemini model showed a clear preference for outputs generated by the model trained on the \texttt{filtered} dataset. Collectively, these findings underscore the value of quality over quantity: intelligently filtering noisy data can significantly improve model performance and the overall informativeness of the training set. However, the results also reveal a trade-off—over-filtering may reduce data diversity, potentially limiting model generalization. Future work should explore more adaptive filtering strategies that balance alignment, variability, and robustness more effectively.

% To start a new column (but not a new page) and help balance the last-page
% column length use \vfill\pagebreak.
% -------------------------------------------------------------------------
% \vfill
% \pagebreak

% References should be produced using the bibtex program from suitable
% BiBTeX files (here: strings, refs, manuals). The IEEEbib.bst bibliography
% style file from IEEE produces unsorted bibliography list.
% -------------------------------------------------------------------------

% \bibliography{strings,refs}
\printbibliography

\end{document}